\documentclass[accepted]{aaaa2021} % after acceptance, for a revised
\usepackage[british]{babel}

\usepackage{natbib} % has a nice set of citation styles and commands
\usepackage{mathtools} % amsmath with fixes and additions
\usepackage{booktabs} % commands to create good-looking tables
\usepackage{algorithm, algorithmic} % algorithm floats
\usepackage{amsfonts} % for mathbb etc
\usepackage{graphicx} % for including graphics files
\usepackage{bbm} % more blackboard math symbols
\usepackage{enumitem} % to reduce item spacing
\usepackage{subcaption}

\title{The Neural Moving Average Model for Scalable Variational Inference of State Space Models}

\author[1,2]{\href{mailto:Tom Ryder <T.Ryder2@newcastle.ac.uk>?Subject=Neural Moving Averge Model paper}{Thomas Ryder}{}} % Lead author
\author[1]{Dennis Prangle}
\author[1]{Andrew Golightly}
\author[1]{Isaac Matthews}
\affil[1]{%
  School of Mathematics Statistics and Physics\\
  Newcastle University\\
  Newcastle\\
  United Kingdom
}
\affil[2]{%
  Huawei Noah's Ark Lab
}

\begin{document}
\maketitle

\begin{abstract}
Variational inference has had great success in scaling approximate Bayesian inference to big data by exploiting mini-batch training.
To date, however, this strategy has been most applicable to models of independent data.
We propose an extension to state space models of time series data based on a novel generative model for latent temporal states: the neural moving average model.
This permits a subsequence to be sampled without drawing from the entire distribution, enabling training iterations to use mini-batches of the time series at low computational cost.
We illustrate our method on autoregressive, Lotka-Volterra, FitzHugh-Nagumo and stochastic volatility models, achieving accurate parameter estimation in a short time.
\end{abstract}

\section{Introduction} \label{sec:intro}

State space models (SSMs) are a flexible and interpretable model class for sequential data,
popular in areas including engineering \citep{elliott2008hidden}, economics \citep{zeng2013state}, epidemiology \citep{fasiolo2016comparison} and neuroscience \citep{paninski2010new}.
SSMs assume a latent Markov chain $x$ with states $x_1, x_2, \ldots, x_T$,
with data as noisy observations of some or all of these.

Standard inference methods for the parameters, $\theta$, of an SSM
require evaluating or estimating the likelihood under various choices of $\theta$ e.g.~using a Kalman or particle filter \citep{Sarkka:2013}.
Each such evaluation has $O(T)$ cost at best, and
even larger costs may be required to control the variance of likelihood estimates.
These methods can thus be impractically expensive for long time series (e.g.~$T \gg 10^6$), which are increasingly common in applications such as genomics \citep{Foti:2014} and geoscience \citep{Foreman:2017}.

In contrast, for models of independent data, one can estimate the log-likelihood using a short \emph{mini-batch}, at an $O(1)$ cost only.
This allows scalable inference methods based on stochastic gradient optimisation e.g.~maximum likelihood or variational inference.
The latter introduces a family of approximate densities for the latent variables indexed by $\phi$.
One then selects $\phi$ to minimise the Kullback-Leibler divergence from the approximate density to the posterior.

We propose a mini-batch variational inference method for SSMs, for the case of continuous states i.e.~$x_i \in \mathbb{R}^d$.
This requires a family of variational approximations $q(\theta, x; \phi)$
with a crucial \emph{locality} property.
It must be possible to sample a subsequence $(x_i)_{a \leq i \leq b}$, to be used as a mini-batch, from the middle of the $x$ sequence at a $O(1)$ cost.
Much existing work on flexibly modelling sequence data
(e.g.~\citealt{Oord:2016, pmlr-v80-ryder18a, Radev:2020})
uses an \textbf{autoregressive model} for $x$.
Here $x_i$ is generated from some or all $x_j$s with $i<j$, so sampling $x_a$ requires sampling $(x_i)_{i < a}$, and the locality property is not met.

To achieve the locality property we introduce the \textbf{neural moving average (nMA) model}.
This is a generative model for sequence data in which a learnable convolutional neural network (CNN) processes
(1) an underlying sequence of base $N(0,1)$ variables and
(2) the sequence of observed data.
The CNN's receptive field has a limited size, rather than encompassing the entire input sequences.
Therefore a sample from the nMA model is a type of moving average of the input sequences (1) and (2):
$x_i$ is produced from a window of values in the input sequences close to position $i$.
This achieves the locality property.
Also, by viewing the nMA model as a type of normalising flow \citep{Rezende:2015:VIN:3045118.3045281, Papamakarios:2019},
we show later that the mini-batch samples can be used to unbiasedly estimate the log-density of the whole $x$ chain, which is crucial to implement variational inference.

A trade-off for producing the locality property is that, under a nMA model, $x_i$ and $x_j$ are independent for $|i-j|$ sufficiently large
i.e.~if they are far enough apart that their CNN receptive fields do not overlap.
Hence using a nMA as the variational approximation assumes no long-range dependence in the posterior for $x$.
Despite this, we demonstrate that our approach works well in several examples. 
These include various challenging observation regimes:
sparse observation times, partial observation of $x_i$, low observation variance and a large number of observations.  
Our flexible variational family produces good posterior estimates in these examples: at best our variational output is indistinguishable from the true posterior.
%We further demonstrate that we can scale to a challenging example with $T=1,000,000$ states in 180 minutes.

The remainder of our paper is as follows.
Section \ref{sec:background} describes state space models.
Section \ref{sec:localIAFs} reviews relevant material on normalising flows and presents the nMA model.
Section \ref{sec:method} sets out our variational inference method.
The nMA model and the resulting inference algorithm are our novel methodological contribution.
Section \ref{sec:experiments} presents our experiments, and Section \ref{sec:conclusions} gives conclusions and opportunities for future work.
Code for the paper is available at \url{https://github.com/Tom-Ryder/VIforSSMs}.

\paragraph{Related Work}

Bayesian inference for SSMs commonly uses sampling-based Markov chain Monte Carlo (MCMC) methods,
involving repeated use of Kalman or particle filters \citep{Doucet2001, Capp:2010:IHM:1965046, Sarkka:2013}.
As discussed above, these methods typically become expensive for long time series,
with each likelihood estimate requiring an $O(T)$ pass through the data.
A recent $O(1)$ sampling-based scheme using a related strategy to ours for scalable SSM inference is \citet{Aicher:2019}.
This approach uses stochastic gradient MCMC with \emph{buffered} gradient estimates, which are based on running a particle filter on a short subsequence of data.
Like our contribution, this approach neglects long-range dependence.

\cite{Aicher:2019}, in common with several other papers discussed here, requires  an observation for each $x_i$.
However many applications involve missing or sparsely observed data.
Our generative model can be applied to such settings as it learns to impute $x_i$ values between observations.

Several stochastic optimisation variational inference methods for SSMs
have previously been proposed, with different variational families for $x$, including:
a multivariate normal distribution with tridiagonal covariance structure
\citep{Archer:2016},
a recurrent neural network \citep{Krishnan:2017},
an autoregressive distribution \citep{Karl:2017, pmlr-v80-ryder18a},
a particle filter \citep{Hirt:2019}.
However, all of these methods have an $O(T)$ cost for each iteration of training and/or require storing $O(T)$ parameters.
% Archer method could marginalise its normal distribution to avoid cost of training, but still has high memory cost of $O(T)$ parameters, which will take $\geq O(T)$ iterations to learn.

\cite{Foti:2014} also propose an $O(1)$ variational inference method based on mini-batch updates.
They consider \emph{hidden Markov models} -- SSMs with discrete states --
which allows the $x$ posterior to be derived using a forward-backward algorithm \citep{Rabiner:1989}.
This would usually require forward and backward passes over the full dataset, at cost $O(T)$,
but they show that approximating these on short subsequences suffices to perform variational inference.
In contrast our paper explores SSMs with continuous states, where a forward-backward algorithm is not available in general \citep{Briers:2010}.
Another difference is that \cite{Foti:2014} use a variational approximation with independence between $\theta$ and $x$,
while our approach avoids this strong assumption.

Parallel Wavenet \citep{Oord:2018} similarly uses a normalising-flow-based generative model for sequence data.
This incorporates long-range dependence using dilated convolutions,
while we use only short range dependence to allow mini-batch inference.
Our local normalising flow is also similar at a high level to a \emph{masked convolutional generative flow} (MACOW) \citep{Ma2019macow}.
%which was published while we were preparing our paper.
The novelty of our approach is that we develop this idea to allow fast variational inference for time series, while \citet{Ma2019macow} focus on density estimation and sampling for image data.

Finally, \citet{Ward:2019} successfully apply our method to mechanistic models with Gaussian process priors placed on unobserved forcing terms, including a multi-output system using real-world data, and Gaussian process regression using a Poisson observation model. 

\section{State Space Models} \label{sec:background}

\paragraph{Notation} Throughout we use $x_i$ to denote an individual state, $x$ to denote the whole sequence of states and $x_{a{:}b}$ to denote a subsequence $(x_i)_{a \leq i \leq b}$.
We use similar notation for sequences represented by other letters.
More generally we use $a{:}b$ to represent the sequence $(a,a+1,\ldots,b)$.

\subsection{Definition}
A SSM is based on a latent Markov chain $x = x_{1{:}T}$.
% These may represent times $t_i = 0, \Delta t, 2\Delta t, \ldots, T$ for some $\Delta t > 0$.
We focus on the case of continuous states $x_i \in \mathbb{R}^d$.
States evolve through a \emph{transition density}
$
p(x_i | x_{i-1}, \theta),
$
with parameters $\theta \in \mathbb{R}^p$.
We assume the initial state is $x_0(\theta)$, a deterministic function of $\theta$.
(This allows examples with initial state known -- $x_0$ is a constant -- or unknown -- $x_0$ depends on unknown parameters.)
Observations $y_i \in \mathbb{R}^{d_y}$ are available for
$
i \in \mathcal{S} \subseteq 0{:}T
$
following an \emph{observation density}
$
p(y_i | x_i, \theta).
$

In the Bayesian framework, after specifying a prior density $p(\theta)$, interest lies in the posterior density 
\begin{equation}
p(\theta,x | y) \propto
p(\theta,x,y) = p(\theta) \prod_{i=1}^{T}
p(x_i | x_{i-1}, \theta) \prod_{i \in \mathcal{S}} p(y_i | x_i, \theta).
\label{eq:bayes_SS}                    
\end{equation}

\subsection{Discretised Stochastic Differential Equations} \label{sec:SDEs}
One application of SSMs, which we use in our examples, is as discrete approximations to stochastic differential equations (SDEs), as follows:
\begin{equation}\label{eq:SDE}
x_{i+1} = x_i + \alpha(x_i, \theta) \Delta t + \sqrt{\beta(x_i,\theta)\Delta t} \epsilon_i,
\end{equation}
where $\epsilon_i \sim N(0, I_d)$ are independent random vectors.
Here $\alpha$ is a $d$-dimensional \textit{drift vector},
$\beta$ is a $d \times d$ positive-definite \textit{diffusion matrix}
and $\sqrt{\beta}$ denotes its Cholesky factor.
The state $x_i$ approximates the state of the SDE process at time $i \Delta t$.
Taking the limit $\Delta t \to 0$ in an appropriate way recovers the exact SDE \citep{sde, Sarkka:2019}.

\section{The Neural Moving Average Model} \label{sec:localIAFs}

Section \ref{sec:intro} gave an intuitive description of the nMA model.
In this section we present a formal description.
First Section \ref{sec:IAFs} presents background material on inverse autoregressive flows (IAFs).
Then Sections \ref{sec:scalar_nMA}--\ref{sec:multivariate} describe the nMA model as a special case of an IAF.
Section \ref{sec:scalar_nMA} describes the case where $x_i$ (a state of the SSM) is scalar,
and Section \ref{sec:multivariate} extends this to the multivariate case.

%In this section we present a flexible generative distribution capable of approximating $x | \theta, y$.
%In Section \ref{sec:method} we describe its use in variational inference for SSMs.
%Our approach builds on \emph{normalising flows} \citep{Rezende:2015:VIN:3045118.3045281, Papamakarios:2019} and \emph{inverse autoregressive flows} (IAFs) \citep{Kingma:2016} in particular.

\subsection{Inverse Autoregressive Flows} \label{sec:IAFs}

A normalising flow represents a random object $x$ as
$
g_m \circ \ldots g_2 \circ g_1(z)
$: a composition of learnable bijections of a base random object $z$.
Here we suppose $x=x_{1:T}$ and $x_i \in \mathbb{R}$.
(Later we consider $x_i$ as a vector.)
We take $z=z_{1:T}$ as independent $N(0,1)$ variables.
By the standard change of variable result,
the log-density of $x$ is
\begin{equation}
\log q(x) = \sum_{i=1}^T \varphi(z_i) - \sum_{j=1}^m \log |\det J_j|
\end{equation}
where $\varphi$ is the $N(0,1)$ log-density function
and $J_j$ is the Jacobian matrix of transformation $g_j$
given input $g_{j-1} \circ \ldots g_2 \circ g_1(z)$.

The bijections in an IAF are mainly \emph{affine layers},
which transform input $z^{\text{in}}$ to output $z^{\text{out}}$ by
\begin{equation} \label{eq:IAF}
z^{\text{out}}_i = \mu_i(z^{\text{in}}_{1{:}{i-1}})
+ \sigma_i(z^{\text{in}}_{1{:}{i-1}}) z^{\text{in}}_i,
\end{equation}
with $\sigma_i > 0$.
This transformation scales and shifts each $z^{\text{in}}_i$.
The shift and scale shift values, $\mu_i$ and $\sigma_i$, are typically neural network outputs.
An efficient approach is to use a single neural network to output all the $\mu_i, \sigma_i$ values for a particular affine layer.
This network uses \emph{masked dense layers} so that $(\mu_i, \sigma_i)$ depends only on $z^{\text{in}}_{1{:}i-1}$ as required \citep{pmlr-v37-germain15, Kingma:2016, Papamakarios:2017:maf}.
In the resulting IAF each affine layer is based on a different neural network of this form.
We'll refer to this as a \emph{masked IAF}.

The shift and scale functions for $z^{\text{out}}_i$ in \eqref{eq:IAF} have an \emph{autoregressive property}: they depend on $z^{\text{in}}$ only through $z^{\text{in}}_j$ with $j<i$.
Hence the Jacobian matrix of the transformation is diagonal with non-zero entries $\sigma_{1:T}$.
The log-density of an IAF made of $m$ affine layers is
\begin{equation} \label{eq:IAFdensity}
\log q(x) = \sum_{i=1}^T \varphi(z_i) - \sum_{j=1}^m \sum_{i=1}^T \log \sigma^j_i
\end{equation}
where $\sigma^j_i$ is the shift value for the $i$th input to the $j$th affine layer.

IAFs typically alternate between affine layers and \emph{permutation layers},
using order reversing or random permutations.
Such layers have Jacobians with absolute determinant 1.
Thus the log-density calculation is unchanged
(interpreting $j$ in \eqref{eq:IAFdensity} to index the $j$th \emph{affine layer} not the $j$th layer of any type).
The supplement (Section G.1) details methods to restrict the output of a IAF e.g.~to ensure all $x_i$s are positive.

IAFs are flexible and, for small $T$, allow fast sampling %-- e.g.~using GPUs --
and calculation of a sample's log-density.
However they are expensive for large $T$ as large neural networks are needed to map between length $T$ sequences.

\subsection{The Neural Moving Average Model} \label{sec:scalar_nMA}

Our neural moving average (nMA) model reduces the number of weights that IAFs require
by using a CNN to calculate the $\mu_i$ and $\sigma_i$ values in an affine layer.
Thus it can be thought of as a kind of local IAF.
Here we explain the main idea by presenting a version for scalar $x_i$.
Section \ref{sec:multivariate} extends this to the vector $x_i$ case.

To define the nMA model we describe how a single affine layer produces its shift and scale values.
%Could add a figure.
The affine layer uses a CNN with input $z^{\text{in}}$, a vector of length $T$.
Let $h^k$ represent the $k$th hidden layer of the CNN, a matrix of dimension $(T,n_k)$ where $n_k$ is %the number of filters in this layer,
a tuning choice.
The first layer applies a convolution with receptive field length $\ell$.
This is an \emph{off-centre convolution} so that row $i$ of $h^1$ is a transformation of $z^{\text{in}}_{i-\ell{:}i-1}$.
We use zero-padding by taking $z^{\text{in}}_i=0$ for $i<0$.
The following hidden layers are length-$1$ convolutions, so row $i$ of $h^{k+1}$ is a transformation of row $i$ of $h^k$.
The output, $h^n$, is a matrix of dimension $(T, 2)$ whose $i$th row contains $\mu_i$ and $\sigma_i$.
The final layer applies a softplus activation to produce the $\sigma_i$ values, ensuring they are positive.
An identity activation is used to produce the $\mu_i$ values.
The $\mu_i$ and $\sigma_i$ values are used in \eqref{eq:IAF} to produce the output of the affine layer.

A nMA model composes several affine layers of the form just described.
Some properties of the distribution for the output sequence $x$ are:
\begin{enumerate}[noitemsep, topsep=0pt]
	\item No long-range dependence: $x_i$ and $x_j$ are independent if $|i-j| > m \ell$, where $m$ is the number of affine layers.
	\item Stationary local dependence: the distributions of $x_{i{:}j}$ and $x_{i+a{:}j+a}$ are the same for most choices of $a$. (Subsequences near to the start of $x$ can differ due to zero-padding.)
\end{enumerate}

To improve the flexibility of the nMA model, affine layers can be alternated with order-reversing permutations.
(Random permutations would not be suitable, as they would disrupt our ability to sample subsequences quickly, as described in Section \ref{sec:sampling}.)
Throughout the paper we consider nMA models \emph{without} order reversing permutation layers,
as we found these models already sufficiently flexible for our examples.
(The supplement, Section C, details how to incorporate these layers.)

We relax stationary local dependence by injecting \emph{local side information} to the CNN
i.e.~giving an extra feature vector $s_i$ as input for each position $i$ in the first CNN layer.
We also use \emph{global} side information to allow $x$ to depend on the parameter values $\theta$
i.e.~giving $\theta$ as extra input for every position $i$.
See the supplement (Sections D, E) for details of the side information we use in practice.

\subsection{Multivariate Case} \label{sec:multivariate}

Here we generalise the nMA model to the case where $x_i \in \mathbb{R}^d$.
We now let $z$ be a sequence $z_1, z_2, \ldots, z_T$ of independent random $N(0,I_d)$ vectors.
A nMA affine layer makes the transformation
\begin{equation} \label{eq:multi affine}
z^{\text{out}}_i = \mu_i + \sigma_i \odot z^{\text{in}}_i,
\end{equation}
scaling the vector $z^{\text{in}}_i$ (elementwise multiplication by vector $\sigma_i$) then shifting it (adding vector $\mu_i$).

In the scalar case it was important to allow complex dependencies \emph{between} $z^{\text{out}}_i$ values.
Now we must also allow dependencies \emph{within} each $z^{\text{out}}_i$ vector.
To do so we use \emph{coupling layers} as in \citet{Dinh:2016}.

We use an extra $k$ subscript to denote the $k$th component of a vector e.g.~$z^{\text{in}}_{ik}$.
We select some $a \approx d/2$.
For $k \leq a$, we take $\mu_{ik}=0$ and $\sigma_{ik}=1$, so that $z^{\text{out}}_{ik}=z^{\text{in}}_{ik}$.
For $k > a$, we compute $\mu_{ik}$ and $\sigma_{ik}$ using a CNN,
modifying the scalar case as follows.
Now row $i$ of $h^1$ is a transformation of $z^{\text{in}}_{i-\ell{:}i-1}$
(the $\ell$ vectors preceding $z^{\text{in}}_i$),
and also $z^{\text{in}}_{ik}$ for $k \leq a$
(the part of $z^{\text{in}}_{i}$ not being modified).
The output $h^n$ is now a tensor of dimension $(T,d-a,2)$ containing $\mu_{ik}$ and $\sigma_{ik}$ values for $k>a$.

This affine layer does not transform the first $a$ components of $z^{\text{in}}_i$.
To allow different components to be transformed in each layer,
we permute components between affine layers.
For example, a $d=2$ permutation layer
transforms $z^{\text{in}}$ to $z^{\text{out}}$
by $z^{\text{out}}_{i1} = z^{\text{in}}_{i2}$, $z^{\text{out}}_{i2} = z^{\text{in}}_{i1}$.
The log-density is now
\begin{equation} \label{eq:q density}
\log q(x) = \sum_{i=1}^T \lambda_i,
\quad \lambda_i =
\varphi(z_i) - \sum_{k=1}^d \sum_{j=1}^m \log \sigma^j_{ik},
\end{equation}
where $\varphi$ is the $N(0,I_d)$ log-density function and
$\sigma^j_{ik}$ is the $k$th entry of the shift vector for position $i$ output by the $j$th affine layer.
Decomposing $\log q(x)$ into $\lambda_i$ contributions will be useful in Section \ref{sec:method}.

\subsection{Sampling} \label{sec:sampling}

Sampling from a nMA model is straightforward.
First sample the base random object $z$.
This is a sequence of length $T$ (of scalars or vectors -- the sampling process is similar in either case).
Now apply the IAF's layers to this in turn.
To apply an affine layer, 
pass the input (and any side information) through the layer's CNN to calculate shift and scale values, then apply the affine transformation.
%i.e.~\eqref{eq:IAF} in the scalar case and XYZ in the multvariate case.
The final output is the sampled sequence $x$.
The cost of sampling in this way is $O(T)$.

In the next section, we will often wish to sample a short subsequence $x_{u{:}v}$.
It is possible to do this at $O(1)$ cost with respect to $T$.
Algorithm A in the supplement gives the details.
In brief, the key insight is that $x_{u{:}v}$ only depends on $z$ through $z_{u-m\ell{:}v}$.
Therefore we sample $z_{u-m\ell{:}v}$ and apply the layers to this subsequence.
The output will contain the correct values of $x_{u{:}v}$.

\section{Variational Inference for SSMs} \label{sec:method}

This section describes how we use nMA models to perform variational inference (VI) efficiently for SSMs.
Section \ref{sec:VIbackground} reviews standard details of VI.
See e.g.~\citet{Blei:2017} for more details.
We then present our novel VI derivation involving nMA models in Section \ref{sec:derivation}
and the resulting algorithm in Section \ref{sec:algorithm}.

\subsection{Variational Inference Background} \label{sec:VIbackground}

We wish to infer the joint posterior density $p(\theta, x | y)$.
We introduce a family of approximations indexed by $\phi$, $q(\theta, x; \phi)$.
Optimisation is used to find $\phi$ minimising the Kullback-Leibler divergence $KL[q(\theta, x; \phi) || p(\theta, x | y)]$.
This is equivalent to maximising the ELBO (evidence lower bound) \citep{Jordan1999},
\begin{align}
\mathcal{L}(\phi) &= E_{\theta,x \sim q} [ r(\theta, x, y, \phi) ], \label{eq:ELBO} \\
\text{for} \quad r(\theta, x, y, \phi) &= \log p(\theta, x, y) - \log q(\theta,x;\phi).
\end{align}
Here $r$ is a log-density ratio.
The optimal $q(\theta, x; \phi)$ approximates the posterior density.
It is typically overconcentrated, unless the approximating family is expressive enough to allow particularly close matches to the posterior.

Optimisation for VI can be performed efficiently using the \emph{reparameterisation trick} \citep{journals/corr/KingmaW13, pmlr-v32-rezende14, icml2014c2_titsias14}.
That is, letting $(\theta, x)$ be the output of an invertible deterministic function $g(\varepsilon, \phi)$ for some random variable $\varepsilon$ with a fixed distribution.
Then the ELBO gradient and unbiased Monte Carlo estimate are
\begin{align}
\nabla \mathcal{L}(\phi) &= E_\varepsilon [ \nabla r(\theta, x, y, \phi) ], \label{eq:ELBOgrad} \\
\widehat{\nabla \mathcal{L}}(\phi) &= 
\frac{1}{n} \sum_{j=1}^n [ \nabla r(\theta^{(j)}, x^{(j)}, y, \phi) ], \label{eq:MonteCarloELBO}
\end{align}
where $(\theta^{(j)}, x^{(j)}) = g(\varepsilon^{(j)}, \phi)$ and
$\varepsilon^{(1)}, \ldots, \varepsilon^{(n)}$ are independent $\varepsilon$ samples.
This gradient estimate can be used in stochastic gradient optimisation algorithms.

\subsection{ELBO Derivation} \label{sec:derivation}

Our variational family for the SSM posterior \eqref{eq:bayes_SS} is
\begin{equation} \label{eq:q}
q(\theta, x; \phi) = q(\theta; \phi_{\theta}) q(x | \theta ; \phi_{x}),
\end{equation}
where $\phi = (\phi_\theta, \phi_x)$.
We use a masked IAF for $q(\theta; \phi_\theta)$
and a nMA model for $q(x | \theta ; \phi_{x})$.
For the latter we inject $\theta$ as side information.
See the supplement (Sections D and E) for more details.

The masked IAF maps a base random vector $z_\theta$ to $\theta$ using parameters $\phi_\theta$, as described in Section \ref{sec:IAFs}.
The nMA model maps $\theta$ and a sequence of vectors $z_x$ to $x$ using parameters $\phi_x$ as described in Section \ref{sec:multivariate}.
Hence we have a mapping from $\varepsilon = (z_\theta, z_x)$ to $g(\varepsilon, \phi) = (\theta, x)$,
allowing us to use the reparameterisation trick below.

This section derives a mini-batch optimisation algorithm to train $\phi$
based on sampling short $x$ subsequences,
so that the cost-per-training-iteration is $O(1)$.
The algorithm is applicable for scalar or multivariate $x_i$.
In this presentation we assume that $\mathcal{S} = 0{:}T$ i.e.~there are observations for all $i$ values.
To relax this assumption remove any terms involving $y_i$ for $i \not \in \mathcal{S}$.

For our variational family \eqref{eq:q}, the ELBO is \eqref{eq:ELBO} with
\begin{equation}
r = \log p(\theta,x,y) - \log q(\theta; \phi_\theta) - \log q(x|\theta; \phi_x).
\label{eq:local h}
\end{equation}
Substituting \eqref{eq:bayes_SS} and \eqref{eq:q density} into \eqref{eq:local h} gives
\begin{equation}
\begin{split}
r =
&\log p(\theta) - \log q(\theta; \phi_\theta) + \log p(y_0 | x_0, \theta) + \\
&\sum_{i=1}^T \big\{ \log p(x_i | x_{i-1}, \theta) + \log p(y_i | x_i, \theta) - \lambda_i
\big\}.
\end{split}
\end{equation}
Now introduce batches $B_1, B_2, \ldots, B_b$:
length $M$ sequences of consecutive integers partitioning $1{:}T$.
Draw $\kappa$ uniformly from $1{:}b$.
Then an unbiased estimate of $r$ is
\begin{equation}
\begin{split}
r_\kappa =
&\log p(\theta) - \log q(\theta; \phi_\theta) + \log p(y_0 | x_0, \theta) + \\
&\frac{T}{M} \sum_{i \in B_\kappa}  \big\{ \! \log p(x_i | x_{i-1}, \theta) + \log p(y_i | x_i, \theta) - \lambda_i
\! \big\}.
\end{split} \label{eq:rkappa}
\end{equation}
Hence an unbiased estimate of the ELBO gradient is
\begin{equation} \label{eq:ELBOest}
\widehat{\nabla \mathcal{L}}(\phi) = \frac{1}{n} \sum_{j=1}^n \nabla r_\kappa(\theta^{(j)},x^{(j)},y,\phi).
\end{equation}
where $(\theta^{(j)}, x^{(j)}) = g(\varepsilon^{(j)}, \phi)$
and $\varepsilon^{(1)}, \ldots, \varepsilon^{(n)}$ are independent $\varepsilon$ samples.

\subsection{Optimisation Algorithm} \label{sec:algorithm}

Algorithm \ref{alg:inference} presents our mini-batch training procedure.
Each iteration of Algorithm \ref{alg:inference} involves sampling a subsequence of $x$ values of length $M+1$.
The cost is $O(1)$ with respect to the total length of the sequence $T$.
%(assuming constant $M$)
This compares favourably to the $O(T)$ cost of sampling the entire $x$ sequence.
Discussion of implementation details is given in the supplement (Section E).

\begin{algorithm}[htb]
	\caption{Mini-batch variational inference for state space models}
	\label{alg:inference}
	\begin{algorithmic}[1]
		\STATE Initialise $\phi_\theta, \phi_x$.
		\LOOP
		\STATE Sample a batch $\kappa$ uniformly from $1{:}b$. Let $u$ and $v$ denote the endpoints of $B_\kappa$.
		\STATE Calculate $\widehat{\nabla \mathcal{L}}(\phi)$ from \eqref{eq:ELBOest}, generating the terms in the sum as follows.
		\FOR {$1 \leq j \leq n$}
		\STATE Sample $\theta^{(j)} \sim q(\theta; \phi_\theta)$.
		\STATE Sample\footnotemark{}
$x^{(j)}_{u-1{:}v}$ from $q(x | \theta; \phi_x)$ (unless $u=1$ in which case sample $x_{u{:}v}$), calculating corresponding $\lambda_{u:v}$ values.
		See Algorithm A in the supplement for details.
		\STATE Calculate $\nabla r_\kappa(\theta^{(j)},x^{(j)},y,\phi)$ using automatic differentiation of \eqref{eq:rkappa}.
		\ENDFOR
		\STATE Update $\phi_\theta, \phi_x$ using stochastic gradient optimisation.
		\ENDLOOP
	\end{algorithmic}
\end{algorithm}
\footnotetext{Note this samples $x_{u-1}$, the state immediately \emph{before} the current batch of interest.
This is needed for the $p(x_i|x_{i-1}, \theta)$ term in \eqref{eq:rkappa} when $i=u$.}

\section{Experiments} \label{sec:experiments}

Below we apply our method to several examples.
All results were obtained using an NVIDIA Titan XP and an 8 core CPU.
For tuning choices and experimental specifics see the supplement (Sections E, F).
Sections \ref{sec:AR1}--\ref{sec:FHN} use simulated data so the results can be compared to true parameter values, while Section \ref{sec:SV} uses real data.

\subsection{AR(1) Model} \label{sec:AR1}
First we consider the AR(1) model
$x_{i+1} = \theta_1 + \theta_2 x_{i} + \theta_3  \epsilon$,
with $\epsilon \sim N(0, 1)$ and $x_0=10$.
We assume observations $y_i \sim N(x_i, 1)$ for $i \in 0{:}T$,
and independent $N(0,10^2)$ priors on $\theta_1, \theta_2, \log \theta_3$.
%NB OBSERVATION AT $i=0$ GIVES NO INFORMATION AS $x_0$ KNOWN
We use this model to investigate how our method scales with larger $T$, and the effect of receptive field length $\ell$.
To judge the accuracy of our results we compare to near-exact posterior inference using MCMC, as described in the supplement (Section A).

\paragraph{Effect of Observation Sequence Length}
We simulated a synthetic dataset for each of four $T$ values:
$5000, 10000, 50000, 100000$
under true parameter values $\theta = (5.0, 0.5, 3.0)$.
We then inferred $\theta$, fixing the hyperparameters so that the cost per iteration for each setting is constant.

\begin{figure*}[tbph]
		\centering
	\begin{subfigure}{.45\textwidth}
		\centering
		\includegraphics[width=\textwidth]{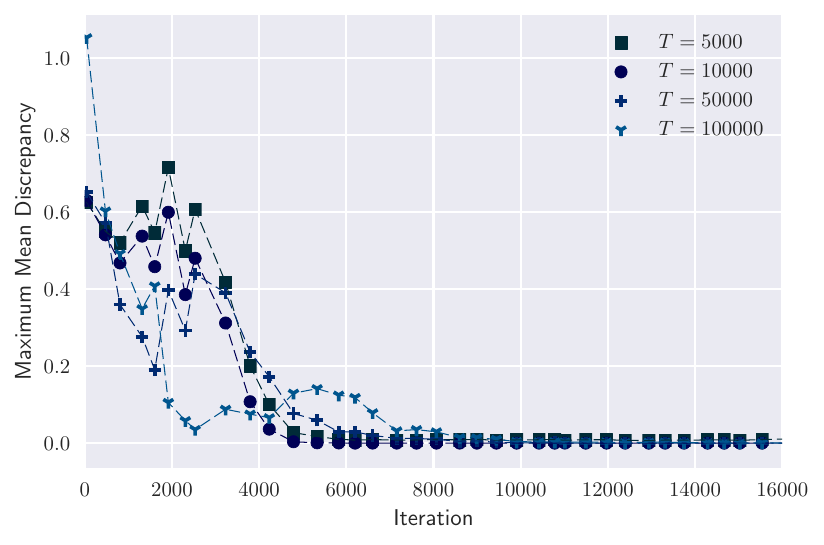}
		\caption{}
		\label{fig:AR_mmd}
	\end{subfigure}	% 
	\begin{subfigure}{.45\textwidth}
		\centering
		\includegraphics[width=\textwidth]{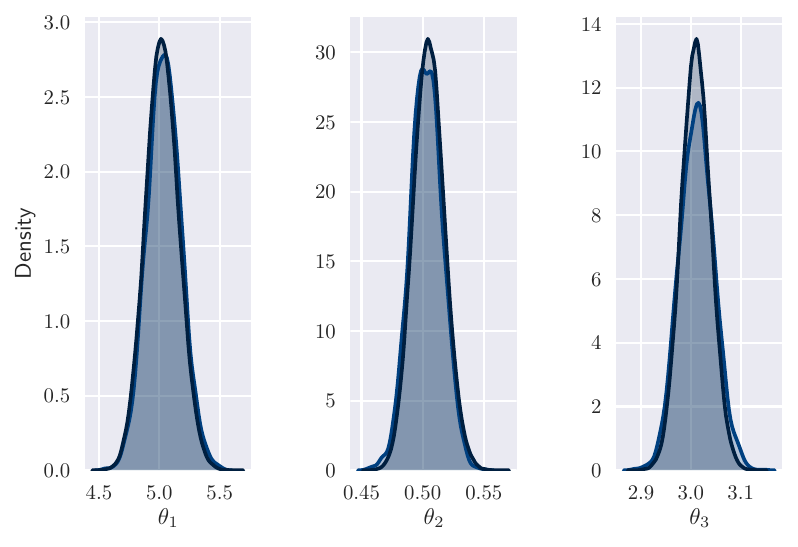}
		\caption{}
		\label{fig:AR_post}
	\end{subfigure}
	\begin{subfigure}{.45\textwidth}
		\centering
		\includegraphics[width=\textwidth]{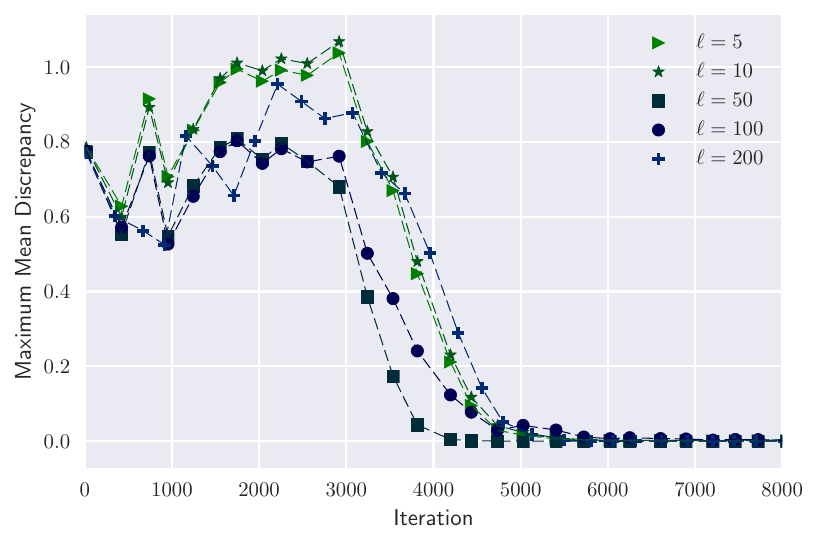}
		\caption{}
		\label{fig:receptive_iter}
	\end{subfigure}%
	\begin{subfigure}{.45\textwidth}
		\centering
		\includegraphics[width=\textwidth]{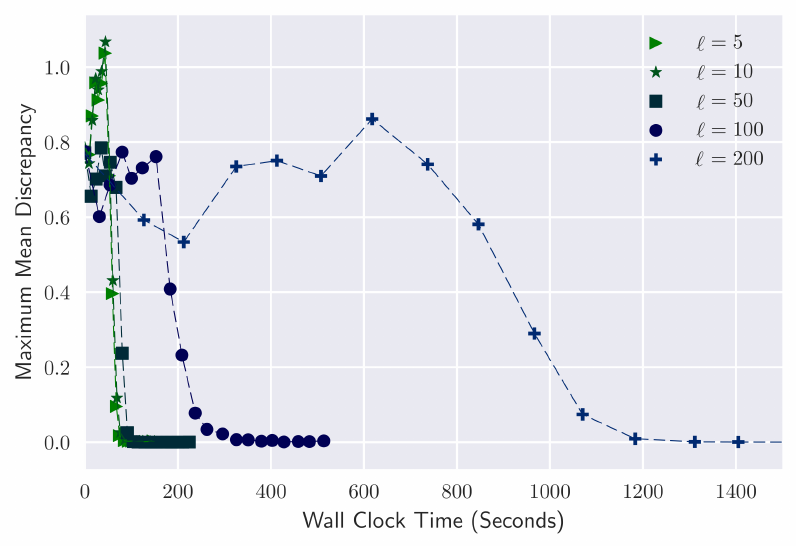}
		\caption{}
		\label{fig:receptive_time}
	\end{subfigure}
	\caption{AR(1) results.
(a) MMD between variational and MCMC output for $\theta$.
(b) Marginal posterior density plots of MCMC output (blue) and variational output after 10,000 iterations (black).
(c,d) MMD between variational and MCMC output for $\theta$ for a range of receptive field lengths $\ell$. The horizontal axis shows (c) number of training iterations (d) wall-clock training time.}
	\label{fig:receptive_choice}
\end{figure*}

Figure \ref*{fig:AR_mmd} plots the accuracy of our results against number of iterations performed.
Accuracy is measured as Maximum Mean Discrepancy (MMD) \citep{Gretton:2012:KTT:2188385.2188410} between variational approximation and MCMC output.
(We use MMD with a Gaussian kernel.)
In all cases, variational inference approximates the posterior well.
Also, the number of training iterations required remains similar as $T$ increases.
As a further check on the quality of the posterior approximation, Figure \ref*{fig:AR_post} shows a good match between marginal posteriors for MCMC and variational output for the case $T=5000$.
Here, as for other $T$ values, the 10,000th iteration is achieved after $\sim 3 $ minutes of computation. 
In comparison, the cost per iteration of MCMC is roughly proportional to $T$.
%For example, we observed a cost roughly 15 times greater for $T=100000$ compared to $T=5000$.

\paragraph{Effect of Receptive Field Length}
We consider again the $T = 5000$ dataset, and investigate the effect of $\ell$.
Figure \ref{fig:receptive_choice} shows MMD against iteration and wall-clock time for $\ell \in \{5, 10, 50, 100, 200\}$. 
In all cases the variational output converges to a good approximation of the posterior.
Convergence takes a similar number of iterations for all choices of $\ell$,
but wall-clock time per iteration increases with $\ell$.

\subsection{Lotka-Volterra}

\begin{figure*}[tbph]
		\centering
	\begin{subfigure}{.45\textwidth}
		\centering
		\includegraphics[width=\textwidth]{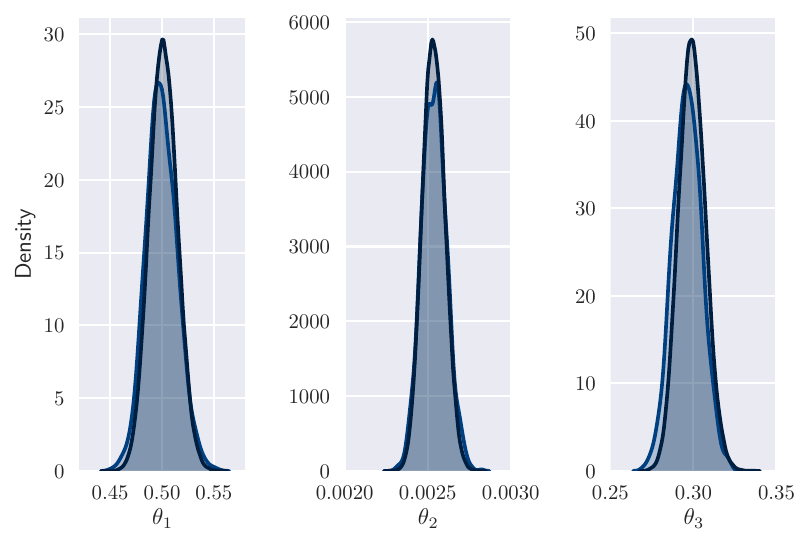}
	\end{subfigure}%
	\begin{subfigure}{.45\textwidth}
		\centering
		\includegraphics[width=\textwidth]{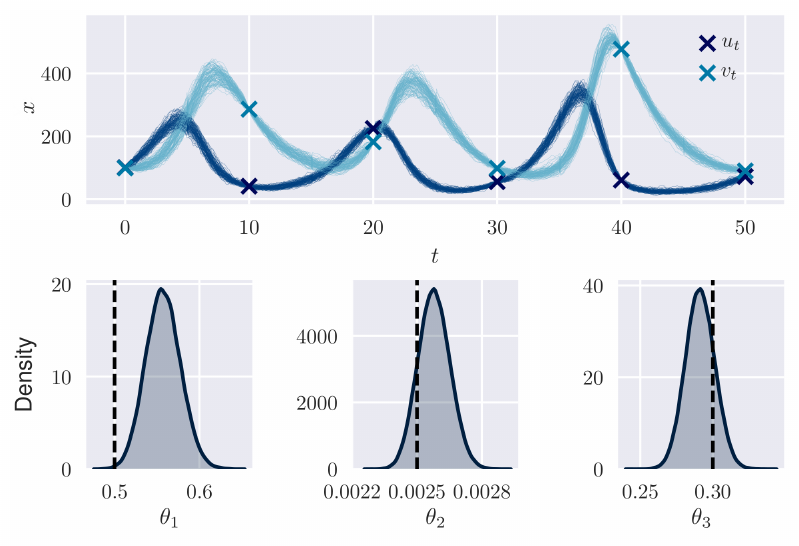}
	\end{subfigure}%
	\caption{Lotka-Volterra results. Left: marginal density plots for setting (a) (dense observations), comparing MCMC (blue) to variational (black) output. Right: variational output for setting (b) (sparse observations). Right top: 100 $x$ samples, with observations displayed as crosses.	The horizontal axis shows $t = 0.1 i$. Right bottom: marginal density plots for $\theta$ with true values displayed with dashed black lines. All variational results used 30,000 training iterations.}
	\label{fig:LV_plots}
\end{figure*}

Next we test our method on short time series with complex dynamics.
We use a version of the Lotka-Volterra model
for predator-prey population dynamics under three events: prey reproduction, predation (in which prey are consumed and predators get resources to reproduce) and predator death.
%I sometimes needed to put an mbox around citation in next sentence to avoid a latex error due to the citation being split over 2 pages.
%See https://tex.stackexchange.com/questions/1522/pdfendlink-ended-up-in-different-nesting-level-than-pdfstartlink.
A SDE Lotka-Volterra model (for derivation see e.g.~\citet{Golightly:2011}) is defined by drift and diffusion

\begin{align}
\alpha(x, \theta) &=
\begin{pmatrix}
\theta_1 u  - \theta_2 u v\\
\theta_2 u v - \theta_3 v
\end{pmatrix}, \label{eq:LVdrift} \\
\beta(x, \theta) &=
\begin{pmatrix}
\theta_1 u  +\theta_2 u v & -\theta_2 u v\\
-\theta_2 u v & \theta_2 u v + \theta_3 v
\end{pmatrix}, \label{eq:LVdiffusion}
\end{align}
where $x = (u, v)$ represents population sizes of prey and predators.
The parameters $\theta = \left(\theta_1, \theta_2, \theta_3\right)$ control the rates of the three events described above.

We consider a discretised version of this SDE, as described in Section \ref{sec:SDEs}, with $\Delta t = 0.1$ and $u_0 = v_0 = 100$.
We simulated realisations under parameters $\theta = (0.5, 0.0025, 0.3)$ of $x_i$ for $i \in 1{:}500$, and construct two datasets with observations at:
(a) $i=0,10,20,\ldots,500$ (dense);
(b) $i=0,100,200,\ldots,500$ (sparse).
%Observations at $i=0$ have no effect as $x_0$ known, so could leave out.
We assume noisy observations $y_i \sim N(x_i, I_2)$
and independent $N(0, 10^2)$ priors for $\log \theta_1, \log \theta_2, \log\theta_3$.

Unlike previous examples, we needed to restrict $u_i$ and $v_i$ to be positive.
%We enforced this by using a final softplus transformation in our flow for $x$.
Also, we found multiple posterior modes, and needed to pretrain carefully to control which we converged to.
See Section G of the supplement for more details of both these issues.

For observation setting (a), we compared our results to near-exact posterior samples from MCMC \citep{Golightly:2008, Fuchs:2013}.
These papers use a Metropolis-within-Gibbs MCMC scheme with carefully chosen proposal constructs.
Designing suitable proposals can be challenging, particularly in sparse observation regimes \citep{Whitaker:2017SC}.
Consequently we were unable to use MCMC in setting (b).

Figure \ref{fig:LV_plots} (left plot) displays the visual similarity between marginal densities estimates from variational and MCMC output in setting (a). The VI output is taken from the 30,000th iteration, after $\approx10$ minutes of computation.  Figure \ref{fig:LV_plots} (right plots) shows variational output for $\theta$ and $x$ in setting (b) after $\approx20$ minutes of computation.
These results are consistent with the ground truth parameter values and $x$ path.
VI using an autoregressive distribution for $x$ has also performed well in a similar scenario \citep{pmlr-v80-ryder18a}, but required more training time (roughly 2 hours).   

\subsection{FitzHugh-Nagumo} \label{sec:FHN}

% Perhaps not the most logical place in the tex file to put the next 2 figures
% but it puts them on the desired page of the output pdf

\begin{figure*}[thb]
	\centering
	\begin{subfigure}{.45\textwidth}
	\centering
	\includegraphics[width=\textwidth]{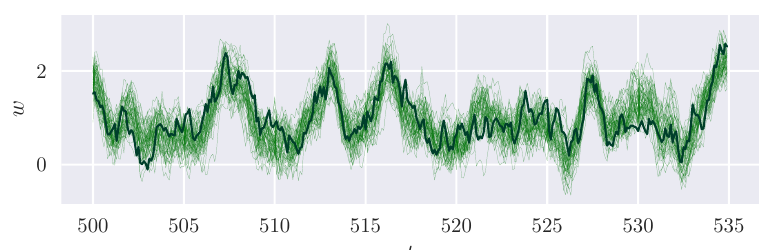}
\end{subfigure}%
\begin{subfigure}{.45\textwidth}
	\centering
	\includegraphics[width=\textwidth]{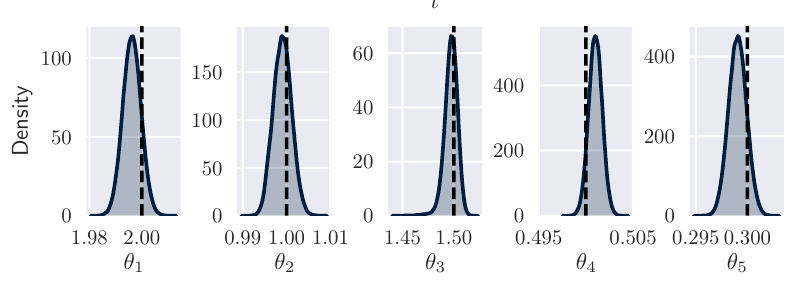}
\end{subfigure}%
\caption{Fitzhugh-Nagumo results.
Left: 100 variational posterior samples (light green) and true values (dark green) for unobserved coordinate $w$ over a short time range.
The $x$-axis shows $t = 0.1 i$.
Right: marginal density plots of variational output for $\theta$ and true values (dashed black lines).}
\label{fig:fhn_out}
\end{figure*}

\begin{figure*}[htb]
	\centering
	\begin{subfigure}{.45\textwidth}
	\centering
	\includegraphics[width=\textwidth]{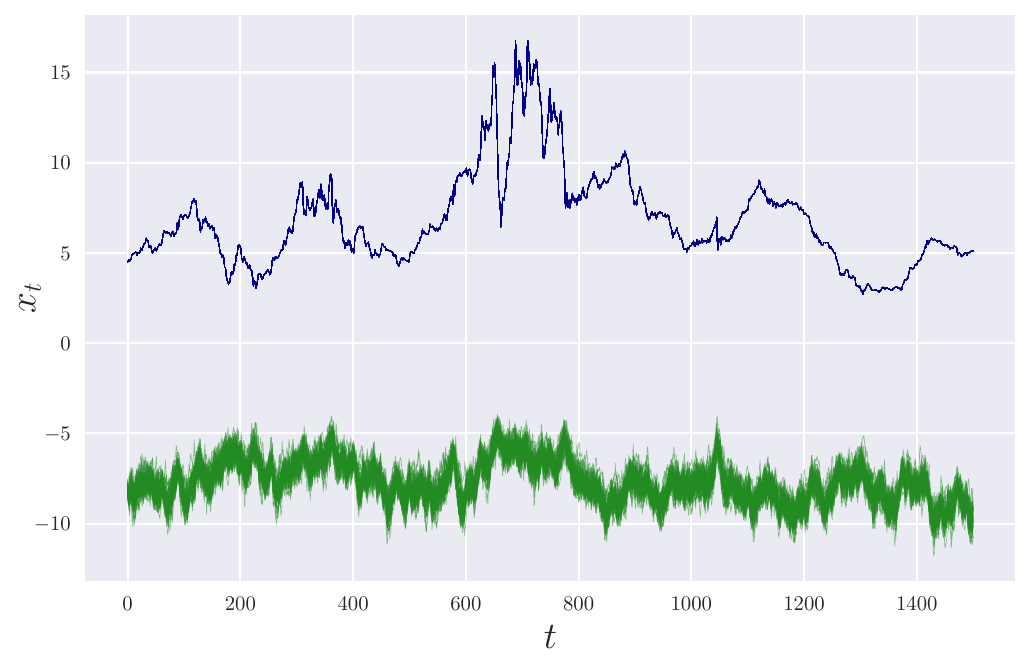}
        \end{subfigure}
	\centering
	\begin{subfigure}{.45\textwidth}
	\centering
	\includegraphics[width=\textwidth]{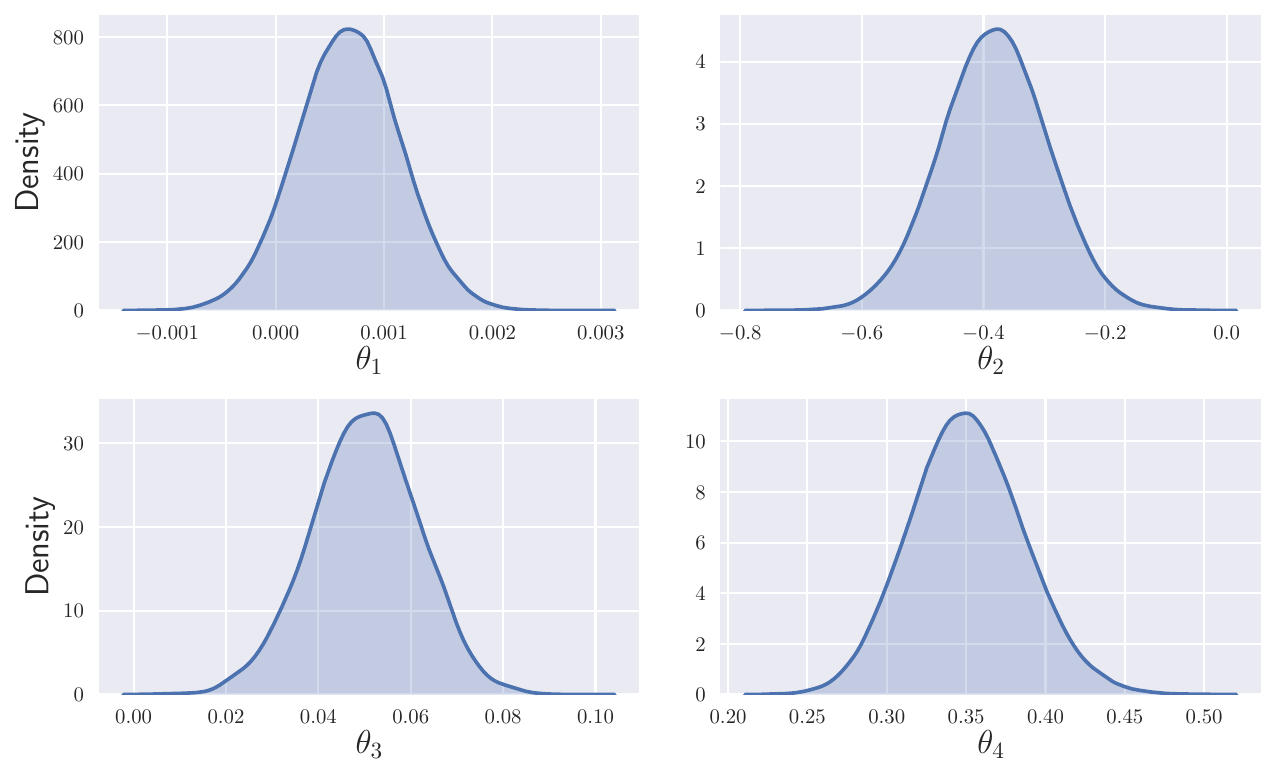}
        \end{subfigure}
	\caption{Stochastic volatility results.
        Left: the observed returns process (blue line) and
        50 samples from the variational posterior for the latent volatility path (green lines).
	Right: Marginal density plots of the variational posterior for $\theta$.}
	\label{fig:SV_results}
\end{figure*}

Here we test our method on a long time series with an unobserved component,
using the FitzHugh-Nagumo model. 
A SDE version, following \citet{Jensen:2012, van2017bayesian}, is defined by drift and diffusion
\begin{align}
\alpha(x, \theta) &=
\begin{pmatrix}
\theta_1 \left(-v^3 + v - w + \theta_2\right)\\
\theta_3 v - w + 1.4
\end{pmatrix}, \\
\beta(x, \theta) &=
\begin{pmatrix}
\theta_4  &0\\
0 & \theta_5
\end{pmatrix},
\end{align}
where $x = (v, w)$ represents the current membrane potential and latent recovery variables.

We consider a discretised version of this SDE, as in Section \ref{sec:SDEs}, with $\Delta t = 0.1, v_0 = 2, w_0 = 3$.
We simulate synthetic data under parameter values $\theta = (2.0,1.0,1.5,0.5, 0.3)$ up to $T=1,000,000$,
recording observations at every $i$
to mimic a high frequency observation scenario.
We assume independent observations $y_i \sim N(v_i, 0.1^2)$
and independent $N(0, 10^2)$ priors for
$\log \theta_1, \theta_2, \theta_3, \log \theta_4, \log \theta_5$.
%Here we challenge our ability to: (1) target the posterior when dealing with an entirely unobserved component; (2) extend inference in a non-trivial model to a longer data setting. 
%
%For this longer time series, we needed to modify our training procedure to avoid large gradient estimates.
%To do so we slowly exposed the model to more of the data over time.
%We began training using only the first 100,000 data points,
%and added an extra 100,000 data points every $1000$ iterations.

Figure \ref{fig:fhn_out} displays estimates of the unobserved component $w$, and marginal density estimates for $\theta$.
The results are consistent with the ground truth parameter values and $w$ path.
The approximate posterior is sampled after roughly $180$ minutes of training.

\subsection{Log-Gaussian Stochastic Volatility} \label{sec:SV}
We analyse a real data under a log-Gaussian stochastic volatility model presented as a discretised SDE with drift and diffusion
\begin{align}
\alpha(x, \theta) &=
\begin{pmatrix}
\theta_1 r\\
\theta_2 - \theta_3 z
\end{pmatrix}, 
& \beta(x, \theta) &=
\begin{pmatrix}
r e^{z} & 0 \\
0 & \theta_4^2
\end{pmatrix},
\end{align}
where $x = \left(r, z\right)$ is the returns process and latent volatility factor, respectively. Similar discrete-time models have been analysed by \citet{lund, Eraker:2001, Durham}, but we use the form presented in \citet{Golightly2006}.

Similarly to \citet{Golightly2006}, we use 1508 weekly observations on the three-month U.S. Treasury bill rate for August 13, 1967 -- August 30, 1996, and perform inference under independent $N(0, 10^2)$ priors for $\theta_1, \theta_2, \log \theta_3, \log \theta_4$ and $z_0$.
We assume the returns process is fully observed without error and set $\Delta t  = 1.0$.
Our analysis took ~20 minutes on a single GPU.
Figure \ref{fig:SV_results} shows the results,
which are consistent with those obtained from the MCMC analysis in \citet{Golightly2006}.

\section{Conclusion} \label{sec:conclusions}

We present a variational inference method for state space models based on a neural moving average model.
This is designed to model complex dependence in the conditional posterior $p(x | \theta, y)$ and be scalable to long time series.
In particular, they allow mini-batch inference where each training iteration has $O(1)$ cost.
We show that our method works well in several applications with challenging features including:
an unobserved state component,
sparse observation times, and
a large number of observations.
Further applications can be found in \citet{Ward:2019}.

Future work could investigate changing several aspects of the flow:
alternating affine transformations with order reversing permutations;
allowing some long-range dependence using a multi-scale architecture;
incorporating recently proposed ideas from the literature \citep{Durkan:2019}.

\begin{contributions} % will be removed in pdf for initial submission,
                      % so you can already fill it to test with the
                      % ‘accepted’ class option
    % Briefly list author contributions.
    % This is a nice way of making clear who did what and to give proper credit.

    % H.~Q.~Bovik conceived the idea and wrote the paper.
    % Coauthor One created the code.
    % Coauthor Two created the figures.
T.~Ryder and D.~Prangle conceived the idea and wrote the paper.
%All authors edited the paper.
T.~Ryder and I.~Matthews wrote the code and performed the experiments.
A.~Golightly advised on SSM models and methods, and wrote Appendix A.
\end{contributions}

\begin{acknowledgements} % will be removed in pdf for initial submission,
                         % so you can already fill it to test with the
                         % ‘accepted’ class option
Thomas Ryder and Isaac Matthews are supported by the Engineering and Physical Sciences Research Council, Centre for Doctoral Training in Cloud Computing for Big Data (grant number EP/L015358/1).

We acknowledge with thanks an NVIDIA academic GPU grant for this project.

We thank Wil Ward and Stephen McGough for helpful discussions,
and anonymous reviewers for useful comments.
\end{acknowledgements}

\bibliography{neuralMA}

\clearpage
\appendix
\setcounter{algorithm}{0}
\renewcommand\thealgorithm{\Alph{algorithm}}

\section*{\qquad Supplementary Material}

\section{MCMC Algorithm for AR(1) via Forward-Filter Recursion } \label{sec:forward_filter}
Here we present an MCMC method for the AR(1) model in the main paper,
\begin{equation} \label{eq:AR1}
x_{i+1} = \theta_1 + \theta_2 x_{i} + \theta_3  \epsilon.
\end{equation}
Recall that we take $\epsilon \sim N(0, 1)$ and $x_0=10$.
We assume observations $y_i \sim N(x_i, 1)$ for $i \in 0{:}T$,
and independent $N(0,10^2)$ priors on $\theta_1, \theta_2, \log \theta_3$.

Assuming $T$ observations, the marginal parameter posterior is given by
\begin{equation}\label{eq:OU_target}
p(\theta|y_{0:T}) \propto p(\theta) p(y_{0:T}|\theta),
\end{equation}
where $p(y_{0:T}|\theta)$ is the marginal likelihood obtained from integrating out the latent variables from $p(\theta, x_{0:T}| y_{0:T})$.
We sample the marginal parameter posterior $p(\theta | y_{0:T})$ using a random walk Metropolis-Hastings scheme.

This appendix describes the key step of evaluating the marginal likelihood given $\theta$, which is achieved using a \emph{forward filter}.
See \citet{West2006} for a general introduction to forward-filtering algorithms for linear state space models.
We adapt this as follows.

As can be seen from \eqref{eq:AR1}, the AR(1) model is linear and Gaussian. Hence, for a Gaussian observation model with variance $\sigma^2$, the marginal likelihood is tractable and can be efficiently computed via a forward-filter recursion. This utilises the factorisation
\begin{equation}
p(y_{0:T} | \theta) = p(y_{0}| \theta) \prod_{i=1}^{T} p(y_{i} | y_{0:i-1}, \theta),
\end{equation}
by recursively evaluating each term.

% Assuming $x_{0} \sim N(a, c)$ a priori, we begin by calculating
% \begin{equation}
% p(y_{0}|\theta) = N(y_{0}; a, c + \sigma^2).
% \end{equation}
% The posterior at $0$ is $x_{0} \big| y_{0}, \theta \sim N(a_0, c_0)$ with
% \begin{align}
% a_0 &= a + c(c+\sigma^2)^{-1}(y_{0}-a),\\
% c_0 &= c - c(c+\sigma^2)^{-1} c.
% \end{align}

Suppose that $x_{i} | y_{0:i} \sim N(a_i, c_i)$.
Since $x_0=10$ we can take $a_0=10, c_0=0$.
It follows that
\begin{align}\label{eq:ff1}
x_{i+1} \big| y_{0:i} \sim N\left(\theta_1 + \theta_2 a_i ,
\theta_2^{2}c_i + \theta_3^2  \right),
\end{align}
which, from the observation model, gives us the one-step-ahead forecast
\begin{align}\label{eq:one_step}
y_{i+1} \big| y_{0:i} \sim N\left(\theta_1 + \theta_2 a_i ,
\theta_2^{2}c_i + \theta_3^2  + \sigma^2 \right).
\end{align}Hence the marginal likelihood can be recursively updated using
\begin{equation}\label{eq:ff2}
p(y_{0:{i+1}}| \theta) = p(y_{0:i} | \theta) p(y_{i+1} | y_{0:i}, \theta),
\end{equation}
where $p(y_{i+1} | y_{0:i}, \theta)$ is the density of \eqref{eq:one_step}.

The next filtering distribution is obtained as $x_{{i+1}} | y_{0:{i+1}} \sim N(a_{i+1}, c_{i+1})$ where
\begin{align}
a_{i+1}=&  \theta_1\! + \theta_2 a_i\! + \frac{\left(\theta_2^{2}c_i + \theta_3^2 \right) \left(y_{i+1} -  \theta_1 - \theta_2 a_i \right)}{\left(\theta_2^{2}c_i + \theta_3^2  + \sigma^2\right)} \label{eq:ff3}\\
c_{i+1} =&  \theta_2^{2}c_i + \theta_3^2  - \frac{\left(\theta_2^{2}c_i + \theta_3^2 \right)^2}{\left(\theta_2^{2}c_i + \theta_3^2  + \sigma^2\right)}.\label{eq:ff4}
\end{align}
Evaluation of \eqref{eq:ff1}-\eqref{eq:ff4} for $i = 0,1,\ldots,T-1$ gives the marginal likelihood $p(y_{0:T} | \theta)$.

\section{Mini-Batch Sampling} \label{sec:minisampling}

Algorithm \ref{alg:IAFsampling} describes how to sample a subsequence $x_{a{:}b}$ from $q(x | \theta; \phi_x)$ without needing to sample the entire $x$ sequence.

Let $z^0$ be the base random sequence and $z^j$ be the sequence after $j$ affine layers.
We assume no layers permuting the sequence order
(but do allow layers permuting the components within each vector in the sequence).
Suppose there are $m$ affine layers, so the output is $x = z^m$.

Algorithm \ref{alg:IAFsampling} presents the multivariate case where $x_i, z^j_i, \mu^j_i, \sigma^j_i$ are all vectors in $\mathbb{R}^d$.
This includes $d=1$ as a special case.
We denote the $k$th entry of $\sigma^j_i$ as $\sigma^i_{ik}$.
Recall that $\varphi$ is the $N(0,I_d)$ log density function.

\begin{algorithm}[htb]
\caption{Sampling a subsequence from a nMA model}
\label{alg:IAFsampling}
\begin{algorithmic}[1]
\STATE Sample $z^0_i$ for $a-c_0 \leq i \leq b$.
These are sampled from independent $N(0,I_d)$ distributions
except when $i \not \in 1{:}T$.
In the latter case $z^0_i$ is a vector of zeros.
\FOR{$1 \leq j \leq m$}
\STATE Apply the CNN with input $z^{j-1}_{a - c_{j-1} {:} b}$.
This outputs $\mu^j_{a - c_j {:} b}$ and $\sigma^j_{a - c_j {:} b}$.
\STATE Calculate $z^j_{a-c_j {:} b}$ using affine transformation $z^j_i = \mu^j_i + \sigma^j_i \odot z^{j-1}_i$.
\STATE Permute components in $z^j$ if necessary.
\ENDFOR
\STATE Return sampled subsequence $x_{a{:}b} = z^m_{a{:}b}$,
and log density contributions $\lambda_{a{:}b}$, where
\[
\lambda_i = \varphi(z^0_i) - \sum_{k=1}^d \sum_{j=1}^m \log \sigma^j_{ik}.
\]
\end{algorithmic}
\end{algorithm}

Each iteration of the algorithm (except the last) must sample $z^j_i$ vectors over an interval of $i$ which is wider than simply $a{:}b$.
The number of extra $z^j_i$s required at the lower end of this interval is
\begin{equation}
c_j = (m - j) \ell.
\end{equation}
In other words, at each iteration the required interval shrinks by $\ell$, the length of the receptive field for $z$.

\section{Order-Reversing Permutations}

The main paper mentions that affine flow layers could be alternated with layers which reverse the order of the $x_{1:T}$ sequence.
This can be accomplished by replacing Algorithm \ref{alg:IAFsampling} above with Algorithm \ref{alg:IAFsampling2}.
We state this algorithm using only the original sequence ordering.
To do so we introduce an operation $\mathfrak{R}$ which reverses the order of a sequence.

As before, each iteration (except the last) samples $z^j_i$ vectors over an interval of $i$ wider than $a{:}b$.
However now we need extra entries at both ends of this interval,
$c^-_j$ and $c^+_j$ at the lower and upper ends respectively.
These can be defined recursively by $c^-_m = c^+_m = 0$ and
\begin{equation}
(c^-_j, c^+_j) = (c^-_{j+1}, c^+_{j+1}) +
\begin{cases}
(\ell, 0) & j \text{ odd} \\
(0, \ell) & j \text{ even}
\end{cases}
\end{equation}

\begin{algorithm}[htb]
\caption{Sampling a subsequence from a nMA model, with order-reversing permutations}
\label{alg:IAFsampling2}
\begin{algorithmic}[1]
\STATE Sample $z^0_i$ for $a-c_0 \leq i \leq b$ as in Algorithm \ref{alg:IAFsampling}.
\FOR{$1 \leq j \leq m$}
\IF{$j$ odd}
\STATE Apply the CNN with input $z^{j-1}_{a - c^-_{j-1} {:} b + c^+_{j-1}}$.
This outputs $\mu^j_{a - c^-_j {:} b + c^+_j}$ and $\sigma^j_{a - c^-_j {:} b + c^+_j}$.
\ELSE
\STATE Apply the CNN with input $\mathfrak{R}(z^{j-1}_{a - c^-_{j-1} {:} b + c^-_{j-1}})$.
This outputs $\mathfrak{R}(\mu^j_{a - c^-_j {:} b + c^+_j})$
and $\mathfrak{R}(\sigma^j_{a - c^-_j {:} b + c^+_j})$.
\ENDIF
\STATE Calculate $z^j_{a - c^-_j {:} b + c^+_j}$ using affine transformation $z^j_i = \mu^j_i + \sigma^j_i \odot z^{j-1}_i$.
\STATE Permute components in $z^j$ if necessary.
\ENDFOR
\STATE Return sampled subsequence $x_{a{:}b} = z^m_{a{:}b}$,
and log density contributions $\lambda_{a{:}b}$, where
\[
\lambda_i = \varphi(z^0_i) - \sum_{k=1}^d \sum_{j=1}^m \log \sigma^j_{ik}.
\]
\end{algorithmic}
\end{algorithm}

\section{Side Information} \label{sec:sideinfo}

Here we give more details of what side information we inject into our nMA model for $x$.
We inject this information into the first layer of the CNN for each of our affine transformations.
Recall that firstly we include the parameters $\theta$ as global side information.
Also we provide local side information,
encoding information in $y$ local to $i$ which is useful for inferring the state $x_i$.

In more detail, first we define $s_i$ to be a vector of data features relevant to $x_i$.
We pick these so that $s_i$ exists for all $i$ even if
(1) no $y_i$ observations exist for $x_i$ or
(2) $i$ is outside the range $0{:}T$.
The data features we use in our examples are listed in the next section.

The side information corresponding to the $i$th position in the sequence processed by the CNN is $\theta$ and the vector $s_{i-\ell'{:}i+\ell'}$.
The tuning parameter $\ell'$ is a receptive field length (like $\ell$ earlier).
This receptive field extends in both directions from the sequence position $i$, so it can take account of both recent and upcoming observations.
The side information is encoded  using a feed-forward network,
and this vector is then used as part of the input to the first layer of the CNN.

\section{Implementation Details for Algorithm 1} \label{sec:impl}

\paragraph{Optimisation}

We use the AdaMax optimiser \citep{DBLP:journals/corr/KingmaB14},
due to its robustness to occasional large gradient estimates.
These sometimes occurred in our training procedure when different batches of the time series had significantly different properties.
See Section \ref{sec:tune_choice} for its tuning choices.
To stabilise optimisation, we also follow \citet{Pascanu:2013:DTR:3042817.3043083} and clip gradients using the global $L_{1}$ norm.

\paragraph{Variational Approximation for $\theta$}

For $q(\theta; \phi_{\theta})$ we use a masked IAF as described in Section 3.1 of the main paper.
In all our examples, this alternates between 5 affine layers and random permutations.
Each affine transformation is based on a masked feed-forward network of 3 layers with 10 hidden units. 

\paragraph{Unequal Batch Sizes}

The main paper assumes the training batch is split into batches $B_1,B_2, \ldots, B_b$ of equal length.
Recall that in this case a batch $B_\kappa$ is sampled at random to use in a training iteration where $\kappa$ is drawn uniformly from $1{:}b$.

Often the length of the data will require batches of unequal lengths to be used.
To do so, simply take $\Pr(\kappa)=|B_\kappa|/T$,
and replace $T/M$ in (15) (in the main paper) with $T/|B_\kappa|$.

\paragraph{Pre-Training}

We found that pre-training our variational approximation to sensible initial values reduced the training time.
A general framework for this is to train $q(\theta; \phi_\theta)$ to be close to the prior,
and $q(x | \theta; \phi_x)$ to be close to the observations, or some other reasonable initial value.
See Section \ref{sec:tune_choice} of for details of how we implemented this in our examples.

One of our examples required more complex pre-training, described below in Section \ref{sec:LVmodes}.
Although our method does sometimes require such non-trivial tuning choices, so do most other competing methods for Bayesian inference of SSMs (see e.g.~\citealp{Sherlock:2015}).

\paragraph{Local Side Information}

Our local side information vector $s_i$ is made up of:
\begin{itemize}[noitemsep, topsep=0pt]
	\item Time $i$.
	\item Binary variable indicating whether or not $i \in \mathcal{S}$ (i.e.~whether there is an observation of $x_i$).
	\item Vector of observations $y_i$ if $i \in \mathcal{S}$. Replaced by the next recorded observation vector if $i \not \in \mathcal{S}$, or by a vector of zeros if there is no next observation. 
	\item Time until next observation (omitted in settings where every $i$ has an observation).
	\item Binary variable indicating whether $i \in 0{:}T$ (as the $s_i$ receptive field can stretch beyond this).
\end{itemize}

\paragraph{Choice of $\ell'$}

Throughout we use $\ell'=10$.
We found that this relatively short receptive field length for local side information was sufficient to give good results for our examples.

\section{Experimental Details}\label{sec:tune_choice}

This section lists tuning choices for our examples.
In all of our examples we set both $n$ (number of samples used in ELBO gradient estimate) and $M$ (batch length) equal to 50, and use $m=3$ affine layers in our flow for $x$.

Each affine layer has a CNN with 4 layers of one-dimensional convolutional networks.
Each intermediate layer has 50 filters, uses ELU activation and batch normalisation (except the output layer).
Before being injected to the first CNN layer,
side information vectors (see Section \ref{sec:sideinfo})
are processed through a feed-forward network to produce an encoded vector of length 50.
We use a vanilla feed-forward network of 50 hidden units by 3 layers, with ELU activation. 

%Additional features of the model -- both $\theta$ and those calculated from the data -- utilise separate networks prior to use by the first convolutional layer as to learn optimal feature embeddings. 

We use the AdaMax optimiser with tuning parameters $\beta_1 = 0.95$ (non-default choice) and $\beta_2  = 0.999$ (default choice).
See the tables below for learning rates used.

Each experiment uses a small number of pre-training SGD iterations for $\phi_\theta$ optimising $E_{\theta \sim q}[p(\theta)]$, the expected prior density.
We separately pre-train $\phi_x$ to optimise an objective detailed in the tables below.
As discussed above (Section \ref{sec:impl}),
where possible we aim to initialise $x$ to be close to the observations, or some other reasonable initial value.

Choices specific to each experiment are listed below.

\subsection*{AR(1)}

\begin{tabular}{|c|p{50mm}| } 
  \hline
  Learning rate & $10^{-3}$                                                          \\ 
  \hline
  Pre-training for $x$  & 500 iterations minimising $E_{\theta, x \sim q}[||x - \hat{y}||_2]$, where $\hat{y}$ is the observed data. \\ 
  \hline
  $\ell$                 & 10                                                                                  \\
  \hline
\end{tabular}

\subsection*{Lotka-Volterra: Data Setting (a)}

\begin{tabular}{|c|p{50mm}| } 
  \hline
  Learning rate & $10^{-3}$ \\ 
  \hline
  Pre-training for $x$ & 500 iterations minimising $E_{\theta, x \sim q}[||x - \hat{y}||_2]$, where $\hat{y}$ is linear interpolation of the data. \\ 
  \hline
  $\ell$ & 20 \\
  \hline
\end{tabular}

\subsection*{Lotka-Volterra: Data Setting (b)}

\begin{tabular}{|c|p{50mm}| } 
  \hline
  Learning rate & $5 \times 10^{-4}$ \\ 
  \hline
  Pre-training for $x$ & 500 iterations maximising $E_{x \sim q}[p(x|\theta^*)]$ where $\theta^* = (0.5, 0.0025, 0.3)$. See Section \ref{sec:LVmodes} for more details. \\
  \hline
  $\ell$ & 20 \\
  \hline
\end{tabular}

\subsection*{FitzHugh-Nagumo}

\begin{tabular}{|c|p{50mm}| } 
  \hline
  Learning rate & $5 \times 10^{-4}$ \\ 
  \hline
  Pre-training for $x$ & 500 iterations minimising $E_{\theta, x \sim q}[||x||_2]$. 
                         Here the model has some unobserved components, so we cannot initialise $x$ close to the observations.
                         Instead we simply encourage $x$ to take small initial values.
  \\
  \hline
  $\ell$ & 20 \\
  \hline
\end{tabular}

\section{Lotka-Volterra Details} \label{sec:LVextra}

Here we discuss some methodology specific to the Lotka-Volterra example in more detail.

\subsection{Restricting \lowercase{$x$} to Positive Values} \label{sec:LVpos}

For our Lotka-Volterra model, $x_i=(u_i,v_i)$ represents two population sizes.
Negative values don't have a natural interpretation, and also cause numerical errors in the model
i.e.~the matrix $\beta$ in (17) may no longer be positive definite so that a Cholesky factor, required in (2), is not available\footnote{
Note all equation references in Section \ref{sec:LVpos} are to the main paper.
}.

Therefore we wish to restrict the support of $q(x|\theta;\phi_x)$ to positive values.
We so by the following method, which can be applied more generally, beyond this specific model.
We add a final elementwise softplus bijection to our nMA model.
Let $\tilde{x}$ be the output before this final bijection.
The log density (7) gains an extra term to become
\begin{equation}
\log q(x) = \sum_{i=1}^T \varphi(z_i) - \sum_{k=1}^d \sum_{j=1}^m \sum_{i=1}^T \log \sigma^j_{ik} - \sum_{k=1}^d \sum_{i=1}^T \gamma(\tilde{x}_{ik}),
\end{equation}
where $\gamma$ is the derivative of the softplus function (i.e.~the logistic function).
The ELBO calculations remain unchanged except for taking
\begin{equation}
\lambda_i = \varphi(z_i) - \sum_{k=1}^d \sum_{j=1}^m \log \sigma^j_{ik} - \sum_{k=1}^d \gamma(\tilde{x}_{ik}).
\end{equation}
We implement our method as before with this modification to $\lambda_i$.

\subsection{Multiple Modes and Pre-Training} \label{sec:LVmodes}
\begin{figure*}[htb]
	\centering
	\includegraphics[width=0.9\textwidth]{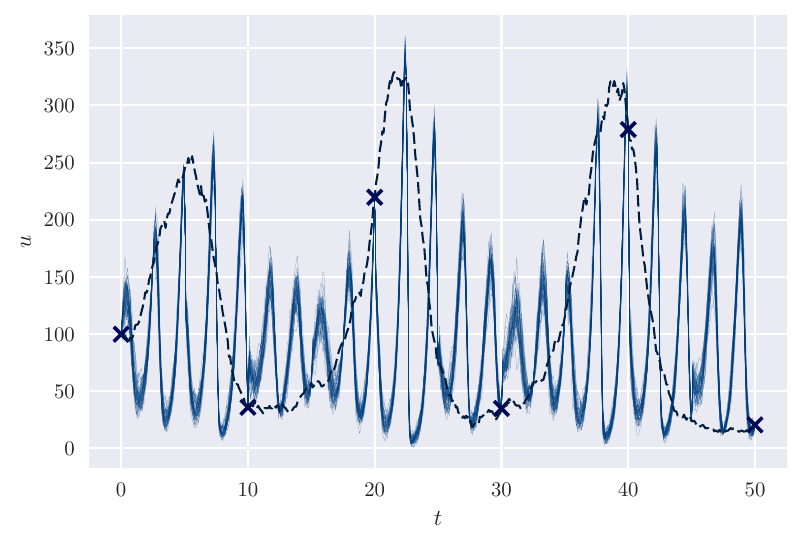}
	\caption{Lotka-Volterra results finding a high-frequency mode. This shows the latent path (dashed line), available observations (crosses) and 50 samples of the variational posterior for $x$. Here, for ease of presentation, we present results for $u$ only.
The horizontal axis shows $t = 0.1 i$.}
	\label{fig:LV_wrong_mode}
\end{figure*}

Observation setting (b) of our Lotka-Volterra example has multiple posterior modes.
Without careful initialisation of $q(x|\theta)$, the variational approach typically finds a mode with high frequency oscillations in $x$.
An example is displayed in Figure \ref{fig:LV_wrong_mode}. 
The corresponding estimated maximum a-posteriori parameter values are $\hat{\theta} = (4.428, 0.029, 2.957)$.

Ideally we would aim to find the most likely modes and evaluate their posterior probabilities,
but this is infeasible for our method.
(It could be feasible to design a reversible jump MCMC algorithm, following \citealp{Green:1995}, to do this,
but we are unaware of such a method for this application.)
Instead we attempt to constrain our analysis to find
the mode we expect to be most plausible -- that giving a single oscillation between each pair of data points.
It is difficult to encode this belief in our prior distribution,
so instead we use pretraining so that VI concentrates on this mode.
This is comparable to the common MCMC tuning strategy of choosing a plausible initial value.

We use 500 pretraining iterations maximising the likelihood of $p(x|\theta^*)$,
where $\theta^* = 0.1 \hat{\theta}$.
The basis for this choice is that periodic Lotka-Volterra dynamics roughly correspond to cycles in $(u,v)$ space.
Multiplying $\theta$, the rate constants of the dynamics, by $\xi$ should give similar dynamics but increase the frequency by a factor of $\xi$.
Based on Figure \ref{fig:LV_wrong_mode} we wish to reduce the frequency by a factor of 10, so we choose $\xi=0.1$.
Using this pre-training approach, we obtain the results shown in the main paper (Figure 3),
corresponding to a more plausible mode.

\end{document}